\documentclass{article}

\usepackage[preprint]{neurips_2025}

\usepackage{caption}
\usepackage[utf8]{inputenc} 
\usepackage[T1]{fontenc}    
\usepackage{hyperref}       
\usepackage{url}            
\usepackage{booktabs}       
\usepackage{amsfonts}       
\usepackage{nicefrac}       
\usepackage{microtype}      
\usepackage{xcolor}         
\usepackage{graphicx}
\usepackage{amsmath}

\usepackage{array}  
\usepackage{multirow}
\usepackage{makecell}
\usepackage[normalem]{ulem}
\useunder{\uline}{\ul}{}
\usepackage{ulem}

\definecolor{background}{RGB}{245,245,245}
\definecolor{string}{RGB}{163,21,21}
\definecolor{keyword}{RGB}{0,0,255}
\definecolor{number}{RGB}{0,128,0}
\usepackage{listings}
\title{Retrieval-augmented Large Language Models for Financial Time Series Forecasting}

\author{%
Mengxi Xiao$^a$, Zhengyu Chen$^b$, Lingfei Qian$^c$, Zihao Jiang$^b$, Yueru He$^c$, \\
\bf
Yijing Xu$^d$, Yuechen Jiang$^e$, Dong Li$^b$, Ruey-Ling Weng$^c$, Jimin Huang$^f$, 
\\ \bf
Min Peng$^a$, Sophia Ananiadou$^f$, Jian-Yun Nie$^g$, Qianqian Xie$^a*$\footnote{Corresponding author.} \\
$^a$ School of Artificial Intelligence, Wuhan University, \\
$^b$ School of Computer Science, Wuhan University, \\
$^c$ The Fin AI,
$^d$ Columbia University, \\
$^e$ Stevens Institute of Technology,
$^f$ University of Manchester, \\
$^g$ University of Montreal
}

\begin{document}

\maketitle

\begin{abstract}
Accurately forecasting stock price movements is critical for informed financial decision-making, supporting applications ranging from algorithmic trading to risk management. However, this task remains challenging due to the difficulty of retrieving subtle yet high-impact patterns from noisy financial time-series data, where conventional retrieval methods, whether based on generic language models or simplistic numeric similarity, often fail to capture the intricate temporal dependencies and context-specific signals essential for precise market prediction.
To bridge this gap, we introduce \textbf{FinSrag}\footnote{Code and data are available at 
\href{https://github.com/The-FinAI/FinSeer}{FinSrag}
.}, the first retrieval-augmented generation (RAG) framework with a novel domain-specific retriever \textbf{FinSeer} for financial time-series forecasting. 
FinSeer leverages a candidate selection mechanism refined by LLM feedback and a similarity-driven training objective to align queries with historically influential sequences while filtering out financial noise. 
Such training enables FinSeer to identify the most relevant time-series data segments for downstream forecasting tasks, unlike embedding or distance-based retrieval methods used in existing RAG frameworks.
The retrieved patterns are then fed into StockLLM, a 1B-parameter LLM fine-tuned for stock movement prediction, which serves as the generative backbone. Beyond the retrieval method, we enrich the retrieval corpus by curating new datasets that integrate a broader set of financial indicators, capturing previously overlooked market dynamics. 
Experiments demonstrate that FinSeer outperforms existing textual retrievers and traditional distance-based retrieval approaches in enhancing the prediction accuracy of StockLLM, underscoring the importance of domain-specific retrieval frameworks in handling the complexity of financial time-series data. 
\end{abstract}

\section{Introduction}
\label{sec:introduction}
Financial time-series forecasting is crucial for maintaining market stability and efficiency, with profound implications for investment decisions, risk assessment, and economic policy formulation~\citep{fama2000forecasting}. However, the extreme volatility and nonlinear dynamics of financial markets pose significant analytical challenges, requiring sophisticated approaches to distill actionable signals from complex, noise-laden data streams. Stock movement prediction, which involves determining future price direction (up or down), represents a particularly critical yet demanding task that has attracted substantial research interest~\citep{xie2023wall,xie2024open,xie2024finben}. Although recent LLM-based approaches have shown promise in stock prediction, they predominantly rely on textual data like news and social media~\citep{wang2024llmfactor}, using only the past several days' closing prices as reference while largely ignoring the wealth of information contained in historical time-series patterns~\citep{bustos2020stock}. This oversight creates a critical gap in effectively harnessing temporal financial data for prediction. The challenge is further compounded by the enormous scale and complexity of such data, spanning multiple influencing factors and extended historical contexts. Addressing this requires an intelligent retrieval mechanism capable of efficiently navigating vast time-series datasets to extract and deliver the most relevant patterns, thereby empowering LLMs to generate more accurate and reliable stock movement forecasts.

Despite the critical importance of retrieving relevant historical patterns for stock movement prediction, current embedding-based and distance-based retrieval-augmented generation (RAG) approaches~\citep{joshi2024robust,he2024g,cui2024tabular} face fundamental limitations when applied to financial time-series data. Conventional embedding-based methods struggle because numeric financial sequences often exhibit superficially similar patterns that lack explicit semantic meaning, making it difficult for text-trained retrievers to distinguish meaningful signals from spurious correlations. Meanwhile, distance-based techniques like Dynamic Time Warping (DTW)~\citep{yang2024timerag} prove inadequate as they typically rely on single-variable comparisons (e.g., adjusted closing prices), ignoring the rich contextual information from other financial indicators. This approach limits retrieval to simplistic trend matching and fails to capture deeper, context-dependent relationships where sequences with divergent or opposing trends may actually contain complementary predictive signals. The inherent complexity of financial markets demands retrieval methods that can move beyond surface-level pattern matching to identify truly informative historical sequences that may exhibit complex, non-obvious relationships to current market conditions.

\begin{figure}[t]
    \centering
    \includegraphics[width=\linewidth]{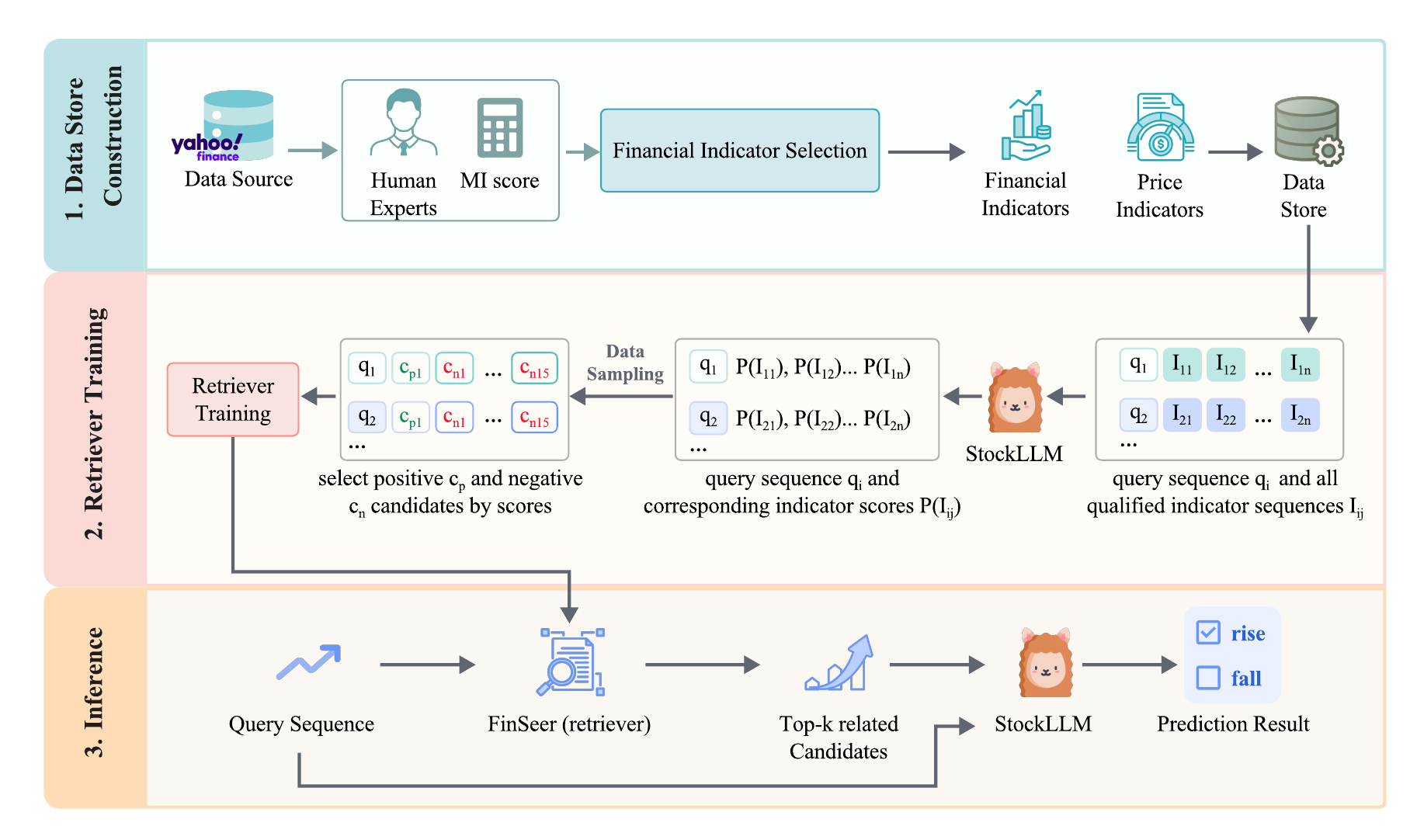}
    \caption{Overview of our time-series RAG framework.}
    \label{fig:framework}
\end{figure}

To address these challenges, we introduce \textbf{FinSrag}, the first \textbf{Fin}ancial time-\textbf{S}eries \textbf{RAG} framework for stock movement prediction, using \textbf{FinSeer}, a \textbf{Fin}ancial Time-\textbf{Se}ries Retriev\textbf{er} to effectively retrieve the most beneficial historical sequences beyond surface-level pattern matching for downstream forecasting.
FinSeer is trained to identify sequences that are embedded with essential or complementary information with limited similarity compared with the query that are beneficial for prediction. A key challenge in training FinSeer lies in the absence of explicit retrieval ground truth, that for any given query, it is inherently ambiguous which historical patterns will most effectively aid forecasting given the vast amount of data. To address this, we propose an LLM-guided relevance estimation, where the language model itself is used to assess and refine the selection of candidate sequences during training.
Building on the approach of~\citet{zhang2023retrieve}, we assess candidate relevance by feeding query-candidate pairs into the LLM and using the generation logits corresponding to the correct answer as a proxy for relevance. Intuitively, candidates yielding higher logits are considered more informative for the forecasting task.

We then train FinSeer to effectively distinguish high-value sequences (positives) from irrelevant sequences (negatives) by learning a retrieval embedding space where relevant candidates are closely aligned with their corresponding queries. Specifically, for each query, FinSeer is optimized to maximize the similarity score between the query embedding and the embedding of top-ranked candidates (i.e., those associated with higher generation logits from the LLM), while simultaneously pushing apart embeddings of low-ranked, uninformative candidates. This contrastive learning objective encourages FinSeer to internalize the LLM’s implicit relevance signals, thereby enhancing its ability to retrieve sequences that are truly beneficial for downstream forecasting.



Beyond improving relevant sequence retrieval, our framework introduces a crucial advancement in what to retrieve by expanding the candidate pool beyond traditional price-based segments. Specifically, we segment the financial time series into candidates, where each candidate corresponds to a time-series segment of a single feature, such as the adjusted close price or a specific technical indicator.
Unlike prior approaches that retrieve only from price sequences, for the first time, our method incorporates 28 additional financial indicators, allowing retrieval over a richer set of candidate sequences that capture diverse aspects of market behavior\footnote{For instance, candidates derived from the KDJ indicator group, particularly in the overbought region where \(K > 80\), \(D > 70\), and \(J > 90\) can reveal early signals of potential reversals~\citep{wu2015technical}.}. To support this, we construct a new dataset and enriched existing two where time-series segments from both price data and a broad set of financial indicators are included as retrieval candidates, enabling FinSeer to discover deeper predictive patterns beyond surface-level price trends.
Comprehensive evaluations across multiple financial time-series datasets demonstrate that FinSeer, our specialized retriever, is the only approach that consistently enhances LLM forecasting accuracy, outperforming five state-of-the-art retrievers from the MTEB leaderboard when integrated into the FinSrag framework. This superior performance not only validates our domain-specific design for time-series data, but also crucially reveals FinSeer's unique capability to identify and retrieve the most predictive financial indicator segments. Notably, the distance-based method which cannot leverage FinSrag's architecture, shows weaker performance, further confirming the advantages of our framework's structured approach to sequence selection, by incorporating diverse financial indicators as candidates.

More than just improving accuracy, FinSeer's retrieval patterns provide valuable financial insights: By automatically discovering and leveraging latent market factors that conventional approaches miss, our method offers an empirically grounded alternative to theoretical assumptions in economic models. This data-driven approach enables systematic identification of predictive market features, opening new avenues to study market inefficiencies and dynamic dependencies through the lens of actual predictive relevance rather than ex-ante hypotheses.
In conclusion, our contributions are summarized as follows:
\begin{enumerate}
\item We introduce FinSrag, the first RAG framework with domain-specific retriever for financial time-series forecasting. By improving the capability of identifying and retrieving the most relevant time-series data segments, FinSrag significantly enhances LLMs' forecasting capabilities in financial markets.

\item We propose FinSeer, a novel domain-specific retriever trained via LLM feedback to uncover overlooked market signals. We complement this with an enriched dataset featuring 28 expert-selected technical indicators that go beyond conventional price data.

\item Experimental results demonstrate the effectiveness of our framework, which surpasseses state-of-the-art retrievers and distance-based methods.
\end{enumerate}

\section{Prelimilaries}
Retrieval-augmented financial time-series forecasting~\citep{sezer2020financial} involves predicting future values or trends ($G$) based on a given query sequence ($q$) and a set of retrieved historical sequences ($c$). These sequences are collected over time at regular intervals. In the retrieval process, the goal of the retrieval model ($R$) is to efficiently identify and extract the most useful historical sequences from a large pool of candidates. By providing relevant context, the retrieval model enhances the forecasting model's ability to make accurate and reliable predictions.
In the specific task of stock movement prediction~\citep{xu2018stock}, the problem is framed as a binary classification task: predicting whether a stock's price will \textit{rise} or \textit{fall} on the next trading day. Given a query sequence \(q\), which represents the stock's price over the previous \(t\) days, the model retrieves relevant sequences as context and predicts the stock's movement \(M_{q,d}\) for the next trading day \(d\). Details about threshold settings are illustrated in Appendix~\ref{append:task_definition}.

\section{Methodology}
\subsection{Retrieval Candidate Pool Design and Dataset Construction}

To support our retrieval framework, we introduce a new dataset comprising high-volume stocks from 2022-2023 and enhance two existing datasets with 28 carefully selected financial indicators. Complete dataset statistics are presented in Table~\ref{tab:dataset}, while partitioning details and additional specifications appear in Appendix~\ref{append:dataset_partition}.
The following parts detail our methodology for: (1) stock price data collection and indicator selection (in \textit{Collection of Indicators}), and (2) temporal scope determination and content specification for queries and candidate sequences (in \textit{Query and Candidate Construction}).

\begin{table}[ht]
\small
\centering
\renewcommand{\arraystretch}{1.3}
\caption{Dataset statistics of query and candidate sequence amounts\protect\footnotemark. }
\label{tab:dataset}
\scalebox{0.7}{
\begin{tabular}{lllcrrr}
\hline 
\textbf{Dataset} & \textbf{License} && \textbf{Trading Dates} & \textbf{Train} & \textbf{Valid} & \textbf{Test}
\\ \hline
\multirow{2}{*}{ACL18~\citep{xu2018stock}} 
& \multirow{2}{*}{MIT License}
& query
& 2015.06.03-2015.12.31  
& 3,312
& 440
& 2,912 \\
&& candidate
& 2014.06.02-2015.12.31
& 441,694
& 67,320
& 444,312
\\
\multirow{2}{*}{BIGDATA22~\citep{soun2022accurate}}
& \multirow{2}{*}{Public}
& query
& 2020.04.09-2020.12.31
& 3,229
& 434
& 2,868
\\
&& candidate
& 2019.04.01-2020.12.31
& 328,372
& 44,778
& 328,372
\\
\multirow{2}{*}{STOCK23}
& \multirow{2}{*}{MIT License}
& query
& 2023.01.03-2023.12.31
& 4,268
& 570
& 4,128
\\
&& candidate
& 2022.01.03-2023.12.31
& 404,736
& 50,592
& 404,736
\\ \hline
\end{tabular}
}
\end{table}

\begin{figure}[h]
    \centering
    \includegraphics[width=\linewidth]{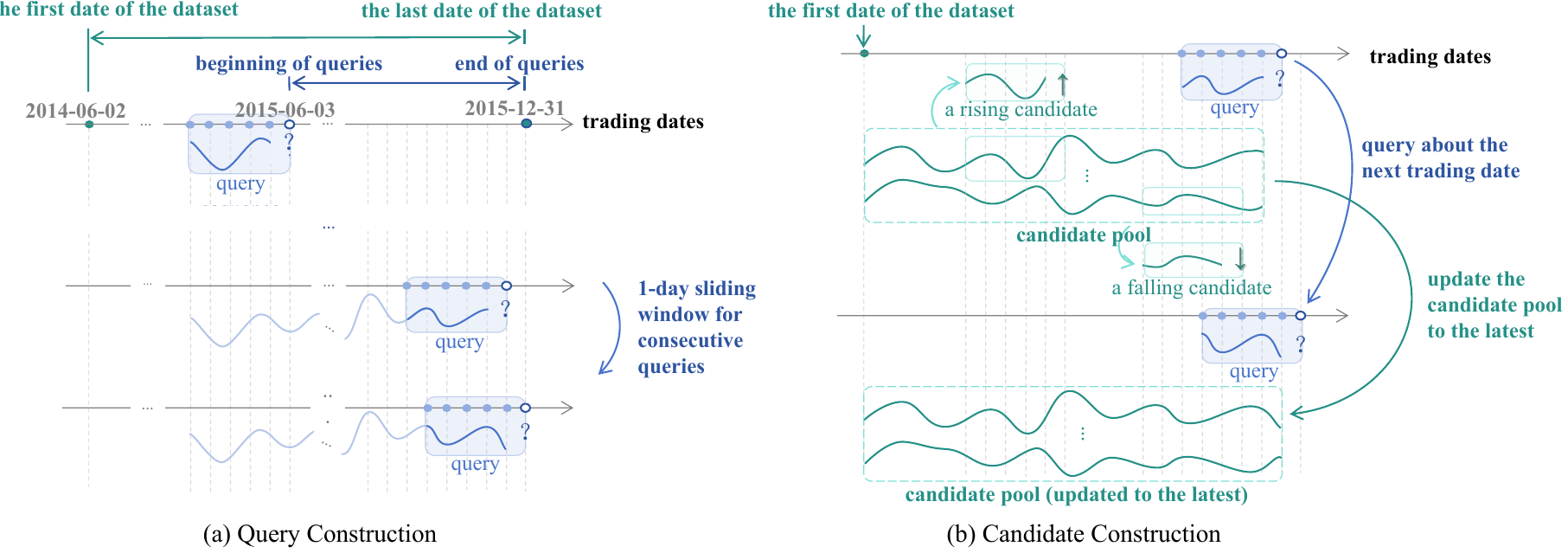}
    \caption{Illustration of query and candidate construction. (a) illustrates how to construct all queries for a given dataset.  (b) illustrates the corresponding candidate pool for a given query, and the update of candidate pool with trading date updates.}
    \label{fig:query_candidate_construction}
\end{figure}

\footnotetext{In the table, the \textit{candidate} rows represent the complete temporal range of each dataset. For any specific query, the corresponding candidate pool spans from the dataset's start date through the trading day immediately preceding the query date, as the example shown in Figure~\ref{fig:query_candidate_construction}(b).}

\noindent
\textbf{Collection of Indicators.}  
Our financial data foundation comes from the Yahoo Finance API~\citep{xu2014stock}, which provides 6 daily trading metrics, including opening, highest, lowest, close prices, adjusted close prices, along with trading volume. While these basic price data offer essential market snapshots, they fail to capture deeper market dynamics crucial for forecasting. For example, if the adjusted close price increases on continuous trading dates, it only shows upward momentum but cannot determine when a trend is exhausted. The KDJ indicator, however, combines price momentum and range position to signal overbought (>80) or oversold (<20) conditions~\cite{wu2015technical}.


\textcolor{black}{These limitations motivate our multi-layer dataset design to extract deeper and latent market signals beyond raw price movements. This multi-layer feature design is inspired by established practices in quantitative finance and machine learning ~\cite{heaton2017deep, jansen2020machine, chen2024deep}, which emphasize the importance of hierarchical signal extraction from raw market data to enhance model predictive power. The dataset consists of three hierarchical components: (1) \textbf{Core Price Metrics}: We begin with aforementioned 6 fundamental indicators, including raw OHLCV (Open, High, Low, Close, Volume) and Adjusted Close prices. (2) \textbf{Primary Technical Indicators}: To capture mid-level market dynamics (e.g., trend strength, volatility, and price-volume interactions) \cite{Lo2000,Park2007,wu2015technical}, which are not directly observable in raw OHLCV, we incorporate 10 widely-used technical indicators such as log returns, price momentum, and VWAP, whose practice is supported by both empirical studies and industry applications \cite{chan2013algorithmic,Fischer2018,Gu2020,Ma2022}. (3) \textbf{Alpha Factors}: To uncover higher-order predictive signals, we identify 18 Alpha Factors via a Mutual Information (MI)-based selection process \cite{guyon2003introduction,jansen2020machine}. This approach quantifies the nonlinear dependency between candidate features and future returns, allowing us to retain only those indicators with strong predictive relationships. The effectiveness of alpha factors in enhancing financial model performance has been extensively supported in the literature \cite{harvey2016and,tulchinsky2019finding,gu2020empirical,kakushadze2016101}.}

The integrated dataset, exemplified in Appendix \ref{sec:datastore_example}, combines these three layers (6 price metrics + 10 technical indicators + 18 alpha factors) to provide a multidimensional view of market conditions. Complete specifications for all 34 indicators appear in Table \ref{tab:indicator_description}.

\textbf{Query and Candidate Construction.}
The query construction process involves temporal range determination and query content specification. For temporal boundaries, we establish that each query must have sufficient candidate sequences available for retrieval by requiring that the query date occurs at least one year after the start date of the corresponding dataset split. This temporal buffer ensures adequate historical context for meaningful retrieval. We implement a one-day sliding window across trading days to generate consecutive queries. For the example shown in Figure~\ref{fig:query_candidate_construction}(a), in the ACL18 dataset spanning from 2014-06-02 to 2015-12-31, we define queries as those occurring between 2015-06-03 and 2015-12-31. This configuration guarantees that even the earliest query (2015-06-03) can access a full year of preceding candidate sequences.
Each query consists of the stock identifier, query date, and recent market data for query-candidate matching. The market data includes adjusted closing prices from the five most recent trading days. The five-day window aligns with standard financial practices, where multiples of five are commonly used for price change calculations. 

The candidate construction follows similar temporal and content specifications as queries. Candidates are drawn from all available historical data preceding each query date, spanning from the dataset's start date to the trading day before the query. We apply a one-day sliding window to generate sequential candidates, with all stocks in the dataset eligible as candidates regardless of query stock matching.
Each candidate consists of the stock ticker, price movement direction on a specific trading date, a relevant financial indicator, and corresponding indicator values calculated from the preceding five consecutive trading days. Figure~\ref{fig:query_candidate_construction}(b) demonstrates this construction through examples of both rising and falling candidates retrieved from the pool.

Our approach enables seamless market monitoring through incremental updates to the candidate pool. When advancing the query date (e.g., from 2015-12-30 to 2015-12-31), we simply append the new trading day's data (2015-12-30) to the existing pool (2014-06-02 through 2015-12-29), maintaining all historical candidates without requiring model retraining. This efficient update mechanism ensures continuous operation while preserving the complete candidate history.

\subsubsection{Sequence Serialization}
\label{sec:serialization}
Since stock movement prediction depends on the changes in related features rather than their exact values, we serialize stock prices and financial indicators into a time-series format. We use JSON to represent these sequences, as it has been demonstrated to effectively support LLMs in interpreting time-series data~\citep{fang2024large,singha2023tabular,yin2023finpt}.

The following are two examples of a query and a candidate sequence. When inquerying about stock MO on 2015-06-02, the query sequence contains the stock identifier (MO), query date (2015-06-02), and the adjusted close prices of last five trading dates. The serialized sequence is shown below:
\begin{lstlisting}[language=Java]
{ "query_stock": "MO",
  "query_date": "2015-06-02",
  "recent_date_list": ["2015-05-26", "2015-05-27", "2015-05-28", "2015-05-29", "2015-06-01"],
  "adjusted_close_list": [29.669, 29.9872, 29.8657, 29.6227, 29.6227]}
\end{lstlisting}

A potential candidate in the candidate pool represent stock MO on date 2014-07-02, with the highest price as its indicator. The sequence includes the stock ticker (MO), price movement direction (freeze) on a specific trading date (2014-07-02), a key financial indicator (the highest price), and corresponding indicator values calculated from the preceding five consecutive trading days. The serialized sequence is shown below:
\begin{lstlisting}[language=Java]
{ "candidate_stock": "MO",
  "candidate_date": "2014-07-02",
  "candidate_movement": "freeze",
  "recent_date_list": ["2014-06-25", "2014-06-26", "2014-06-27", "2014-06-30", "2014-07-01"],
  "high_list": [42.2, 42.0, 41.86, 42.28, 42.0]}
\end{lstlisting}

\subsection{Retriever Training}
\label{sec:training}
We then train FinSeer to effectively distinguish high-value sequences (positives) from irrelevant sequences (negatives) by learning a retrieval embedding space where relevant candidates are closely aligned with their corresponding queries. To achieve this, we score and select positive and negative candidates as the retriever training corpus (in \textit{Candidate Scoring and Selection}), then specify the training objective (in \textit{Training Objective}) and conduct knowledge distillation (in \textit{Knowledge Distillation}).

\noindent
\textbf{Candidate Scoring and Selection.}  
To determine whether a candidate sequence assists in predicting the movement of the query, we use LLM feedback to score each candidate. The details of our LLM backbone, StockLLM-1B-Instruct (hereafter referred to as StockLLM), are shown in Appendix~\ref{append:finetuning_details}. Specifically, for a given query \( q \), we integrate the query sequence and each candidate sequence \( c_i \) from the candidate pool as concurrent inputs to the StockLLM. Then StockLLM outputs logits, which are unnormalized scores representing the model's confidence for each possible class (e.g., \textit{rise} or \textit{fall}). These logits are transformed into probabilities \( P(c) \) using the softmax function:
\begin{equation}
P(c) = \frac{e^{z_c}}{\sum_{j} e^{z_j}},
\end{equation}
where \( z_c \) is the logit for the correct class (e.g., \textit{rise} if the true movement is upward) and \( z_j \) represents the logits for all possible classes. The resulting probability \( P(c) \) serves as the score for the candidate \( c_i \) with respect to the query \( q \).

We rank the candidate sequences in descending order based on their scores \( P(c) \). The top-1 sequence is selected as a positive candidate, while the bottom 15 sequences are chosen as negative candidates. The sets of selected positive and negative sequences are denoted as \( \mathbb{C_P} \) and \( \mathbb{C_N} \), respectively.

\noindent
\textbf{Training Objective.}
Our retriever \( R(q) \) is designed to intelligently distinguish between historically significant sequences \( \mathbb{C_P} \) and noisy sequences \( \mathbb{C_N} \). The training objective is to ensure that \( R(q) \) prioritizes sequences from \( \mathbb{C_P} \) while minimizing attention to those from \( \mathbb{C_N} \). This is achieved by maximizing a similarity measure \( sup(q, s) \) between the query sequence \( q \) and candidate sequences \( s \). Mathematically, the retriever's objective is formulated as:
\begin{equation}
R(q) = {\arg \max}_{s \in \mathbb{C_P} \cup \mathbb{C_N}} sup(q, s).
\end{equation}
By focusing on sequences that maximize \( sup(q, s) \), the retriever ensures that the most informative and contextually relevant historical sequences are identified.

\noindent
\textbf{Knowledge Distillation.}  
To leverage the scoring derived from StockLLM, we employ knowledge distillation, which transfers knowledge from the teacher model (StockLLM) to the student model (retriever) by mimicking the teacher's output distribution. This approach effectively captures nuanced patterns and predictions from StockLLM.  
Specifically, we minimize the Kullback-Leibler (KL) divergence between the candidate distributions computed using the LLM's rewards and those predicted by the embedding model. For each query \( q \) and its candidate list \( \{ \mathbb{C_P}, \mathbb{C_N} \} \), we derive StockLLM's rewards for the candidates, denoted as \( \{P(c_i), i=1, ..., n\} \). To make these rewards suitable for distillation, we normalize them using a softmax function with temperature \( \alpha \):  
\begin{equation}
w_i = \text{softmax}_R \left( \frac{P(c_i)}{\alpha} \right). 
\end{equation}
The KL divergence is then computed as follows:  
\begin{equation}
\min \sum_{c} -w_i \times \log \left( \frac{\exp \left( \left\langle \boldsymbol{e}_q, \boldsymbol{e}_{c_i} \right\rangle / \tau \right)}{\sum_{{c}^{\prime} \in \mathbb{C}} \exp \left( \left\langle \boldsymbol{e}_q, \boldsymbol{e}_{{c}^{\prime}} \right\rangle / \tau \right)} \right),
\end{equation}
where \( \boldsymbol{e}_q \) and \( \boldsymbol{e}_{c_i} \) are the embeddings of the query \( q \) and candidate \( c_i \), respectively, and \( \tau \) is a temperature parameter. This loss function optimizes the similarity between the query embedding and the embeddings of the top-ranked candidates, enhancing the retriever's ability to accurately predict stock price movements.

\subsection{Inference}  
During inference, the key innovation of our FinSrag framework lies in how FinSeer's retrieval directly enhances StockLLM's forecasting capability. Given a query, FinSeer first identifies the most relevant historical sequences by evaluating both temporal patterns and predictive relationships learned during training, filtering out noisy but numerically similar candidates that typically mislead traditional retrievers. These selected sequences are then structured and injected into StockLLM's context window. Crucially, unlike standard RAG that simply concatenates retrieved documents, this end-to-end alignment between retrieval and generation is what enables FinSrag to outperform conventional forecasting pipelines where retrieval and prediction models are optimized separately.

\section{Experiment}
\subsection{Experimental Settings}
\label{section:settings}
\noindent
\textbf{Datasets.}  
We evaluate the effectiveness of our RAG framework on the test sets of the three datasets described in Table~\ref{tab:dataset}, with ACL18 containing 2,876 query sequences, BIGDATA22 containing 2,868 queries, and STOCK23 containing 4,128 queries. These thousand-scale queries help mitigate random bias, ensuring a robust and reliable evaluation of model performance.

\noindent 
\textbf{Candidate Pool Settings.} 
To ensure a comprehensive evaluation, for each query sequence, we include all sequences containing financial indicators across all stocks in the test set (not limited to the same stock), with data available up to the query date, as potential candidates. No additional restrictions are imposed, enabling a robust assessment of the models' performance in real-world financial forecasting scenarios.

\noindent
\textbf{Baselines.}  
To evaluate the efficiency of the FinSrag framework, the bare StockLLM-1B-Instruct without retrieval serves as our baseline to figure out whether the retrieval step enhances StockLLM's prediction ability.
To evaluate our retriever FinSeer, we tested other retrieval methods, including random retrieval, DTW distance, and five competitive retrieving models from the top of the MTEB English Leaderboard as baselines, containing:
(1) Instructor-large~\citep{SuSKWHOYSZ023}, a 335M instruction-finetuned text embedder that encodes sequences into 768-dimensional tensors. 
(2) UAE-large-v1~\citep{li2023angle}, a 335M ANGLE-optimized text embedding model that encodes sequences into 1024-dimensional tensors. 
(3) E5-mistral-7b-instruct~\citep{wang2023improving}, a 7111M embedder initialized from Mistral-7B-v0.1~\citep{jiang2023mistral} and fine-tuned on multilingual datasets, encoding sequences into 4096-di-mensional tensors.  
(4) BGE-large-en-v1.5~\citep{bge_embedding}, a 335M general embedder pre-trained with RetroMAE~\citep{RetroMAE}, encoding sequences into 1024-dimensional tensors.
(5) LLM-Embedder~\citep{zhang2023retrieve}, a 109M embedder fine-tuned from BGE-large-en-v1.5, designed as a unified embedding model to support diverse retrieval augmentation needs for LLMs. It encodes sequences into 768-dimensional tensors. 
More details are shown in Appendix~\ref{append:baseline_settings}.

\noindent
\textbf{Evaluation Metrics.}
We employ Accuracy (ACC) and Matthews
Correlation Coefficient (MCC) \citep{matthews1975comparison} to assess the performance of FinSeer and the baseline models on the stock movement prediction task. These metrics evaluate the performance of stock movement prediction based on the distribution of positive and negative samples. 

\subsection{Main Results}
\label{section:result}
As shown in Table \ref{tab:main-result-1b}, experimental results demonstrate our framework's effectiveness by outperforming all baseline retrieval methods in assisting LLM-based stock movement prediction.
\begin{table}[ht]
\small
\centering
\renewcommand{\arraystretch}{1.3}
\caption{Results of stock movement predictions using StockLLM-1B-Instruct  and retrieval models.}
\label{tab:main-result-1b}
\scalebox{0.85}{
\begin{tabular}{lcccccccc}
\hline
{Retrieving Methods}
& \multicolumn{2}{c}{ACL18} 
&  
& \multicolumn{2}{c}{BIGDATA22} 
&  
& \multicolumn{2}{c}{STOCK23} 
\\ 
\cline{2-3} 
\cline{5-6} 
\cline{8-9} 
(+ StockLLM-1B-Instruct )
& ACC & MCC &  
& ACC & MCC & 
& ACC & MCC 
\\ \hline
w/o Retrieval   
& 0.498 & -0.006 &
& 0.493& -0.017 &
& 0.509 & 0.021     \\
Random Retrieval 
& 0.485 & -0.028 &
& 0.495 & -0.007 &
& 0.496 & -0.004 \\
DTW
& 0.516 & \textbf{0.041} &
& 0.500 & 0.010 &
& 0.492 & -0.007 \\
Instructor~\citep{SuSKWHOYSZ023}     
& 0.498 & -0.005 &
& 0.493 & -0.015 &
& 0.505 & 0.010 \\
UAE~\citep{li2023angle}             
& 0.486 & -0.029 &
& 0.493 & -0.011 &
& 0.494 & -0.009 \\
E5~\citep{wang2023improving}             
& 0.510 &  0.027 &
& 0.498 & -0.002 &
& 0.499 & -0.001 \\
BGE~\citep{bge_embedding}         
& 0.492 & -0.012 &
& 0.501 & 0.002 &
& 0.488 & -0.014 \\
LLM Embedder~\citep{zhang2023retrieve}    
& 0.503 &  0.007 &
& 0.459 & -0.083 &
& 0.503 & 0.007  
\\ \hline
FinSeer        
& \textbf{0.517} & {0.035} &
& \textbf{0.510} & \textbf{0.023} &
& \textbf{0.542} & \textbf{0.085}
\\ \hline
\end{tabular}
}
\end{table}

First, compared with bare StockLLM and random retrieval, FinSeer demonstrates the critical importance of retrieving truly valuable information. While trained on a comprehensive stock movement prediction corpus, bare StockLLM's limited input with only the recent five-day adjusted close price results in unstable performance that fluctuates around random guessing levels. Similarly, when randomly retrieved sequences are provided as supposedly relevant context, they introduce instability by arbitrarily confusing or occasionally coincidentally benefiting StockLLM's decision-making.

Second, our comparison with the five top-ranked retrievers from the MTEB English leaderboard reveals FinSeer's superiority in time-series retrieval.
The negligible performance gap between instruction-finetuned retriever (Instructor) and no-retrieval baselines underscores the fundamental challenges of time-series retrieval. Unlike text retrieval, this task cannot rely solely on task understanding since candidate sequences often exhibit visual similarity while differing in predictive value. 
Other retrievers demonstrate inconsistent cross-dataset performance because their similarity differentiation fails to align with the LLM's perception of importance. Even LLM Embedder, our backbone model trained with LLM feedback, shows limited generalization to time-series retrieval, further emphasizing the problem's complexity.
Among those retrievers, FinSeer consistently outperforms other retrievers across all datasets, proving its superior ability to learn LLM preferences and effectively enhance time-series forecasting through retrieval.

Third, our framework's advantages over distance-based retrieval methods highlight the value of incorporating diverse feature types. While DTW achieves comparable performance to FinSeer during specific market periods (ACL18 [2014-2015] and BIGDATA22 [2019-2020]), its retrieval capability proves insufficient during the volatile 2022-2023 period (STOCK23) marked by significant market surges and fluctuations. During this challenging phase, DTW significantly degrades StockLLM's performance, while FinSeer maintains its enhancement capability.

\subsection{Ablation Study}
\label{sec:ablation}
In this section, we explore two aspects of our framework based on retrieval results: which indicators are most retrieved by all retrievers (in \textit{Indicator Occurrences}), and whether StockLLM-1B-Instruct makes predictions by analyzing candidates or just based on the candidates' movements (in Appendix~\ref{append:candidate_movement_relation}). We also visualize indicator sequence embeddings in Appendix~\ref{append:visualization}. Moreover, we explore the performance of these retrieval methods with a larger size of StockLLM (in Appendix~\ref{append:result-8b}).

\begin{figure*}[t]
\centering
\includegraphics[width=0.9\linewidth]{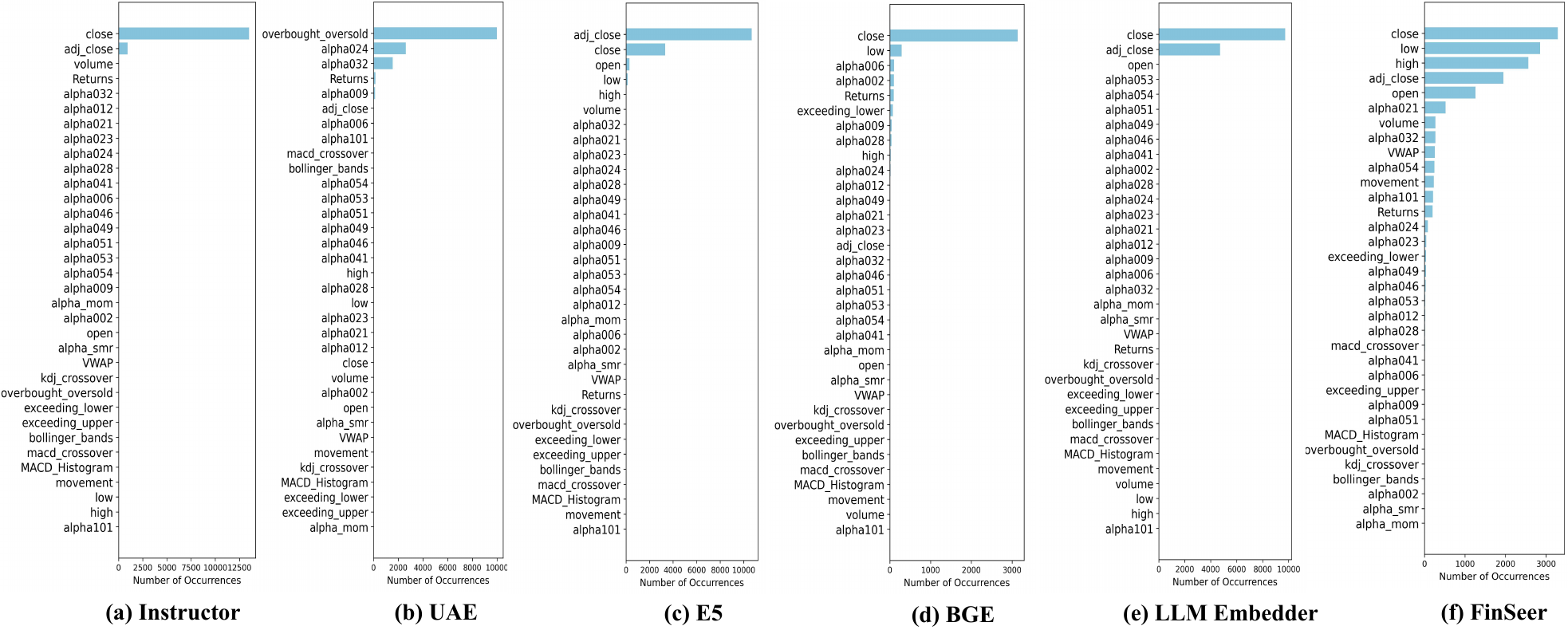}
\caption{Indicator occurrences of different RAG models on ACL18 dataset.}
\label{fig:acl18_indicators}
\end{figure*}

In this section, we analyze the retrieved indicators of all RAG models. We calculate indicator occurrences on the ACL18 test set, and the results are shown in Figure \ref{fig:acl18_indicators}. As shown in the figure, FinSeer is the only model that successfully extracts a diverse and comprehensive set of indicators while achieving superior performance. This clearly demonstrates its advanced temporal retrieval capabilities. Specifically, while other models like LLM Embedder and Instructor predominantly focus on basic indicators such as close price and adjusted close price, FinSeer effectively identifies and retrieves a wide range of technical indicators, including kdj crossover, MACD Histogram, Bollinger Bands, and various alpha factors. This richer set of retrieved indicators provides FinSeer with more comprehensive auxiliary information, enabling more accurate and reliable predictions.

\subsection{Case Study}
This case study illustrates the critical importance of alignment between the retriever and the LLM's forecasting preferences in financial time-series analysis. We examine the stock XOM on 2015-06-25 from the ACL18 dataset, where the adjusted close price exhibited a pronounced downward trend. The query sequence is as follows:
\begin{lstlisting}[language=Java]
{ "query_stock": "XOM",
  "query_date": "2015-06-25",
  "recent_date_list": ["2015-06-18", "2015-06-19", "2015-06-22", "2015-06-23", "2015-06-24"],
  "adjusted_close_list": [58.0813, 57.8979, 57.8707, 57.8027, 57.5377]}
\end{lstlisting}

While multiple retrievers were evaluated, only FinSeer successfully enabled StockLLM to predict the correct movement as a fall. Specifically, FinSeer retrieved five diverse indicators: close price, adjusted close price, alpha021, alpha054, and the highest price, providing a comprehensive view of the stock's behavior. Alpha021 identifies trends based on short- and long-term price averages and volume conditions, while alpha054 combines price and volume rankings to assess performance within a specific time window. These indicators allowed StockLLM to accurately assess whether the downward trend would persist or reverse, demonstrating the value of retrieving contextually relevant and diverse features.

In contrast, other retrievers, such as Instructor, BGE, LLM Embedder, and E5, extracted sequences dominated by close or adjusted close prices, all reflecting similar downward trends. While these sequences aligned with the current trend, they failed to provide actionable insights for forecasting future movements, leading StockLLM to misinterpret them as noise and incorrectly predict a rise. Similarly, UAE retrieved sequences indicating overbought and oversold conditions, including three rise and two freeze trends. Although overbought signals often suggest a potential downturn, the retrieved sequences themselves exhibited rising or frozen trends, confusing StockLLM and resulting in an erroneous prediction. This case study underscores the superiority of FinSeer in retrieving meaningful and diverse indicators that align with the LLM's forecasting logic, enabling more accurate and reliable predictions.

\section{Conclusion}
In this paper, we present FinSrag, the first retrieval-augmented generation framework tailored for financial time-series forecasting. At its core, FinSeer, a novel retriever refined by LLM feedback, effectively identifies historically influential market sequences while filtering out financial noise. Combined with StockLLM, a fine-tuned LLM with 1B parameters, our framework leverages enriched financial datasets to capture previously overlooked market dynamics.
Empirical results confirm that FinSeer surpasses both textual and distance-based retrievers in improving StockLLM’s prediction accuracy, highlighting the necessity of domain-specific retrieval in financial forecasting. Beyond stock markets, FinSrag establishes a blueprint for integrating RAG in time-sensitive decision-making domains.


\bibliographystyle{unsrtnat}
\bibliography{main}

\appendix
\clearpage

\section{Limitation}
\label{section:limitation}
While our FinSrag framework demonstrates significant improvements in LLM-based time-series forecasting, several limitations suggest important directions for future research. First, although our current evaluation focuses specifically on stock movement prediction, the core methodology and architectural principles of our framework are inherently generalizable to other time-series prediction tasks. Future work will systematically validate this generalizability across domains, including but not limited to economic indicators, energy demand forecasting, and epidemiological trend prediction.

Second, the present study deliberately focuses on unimodal time-series data to provide clear, controlled evidence for the effectiveness of our retrieval mechanisms. This focused approach has allowed us to rigorously quantify the contribution of temporal pattern retrieval independent of other factors. However, we recognize that many real-world forecasting scenarios could benefit from multimodal integration, where time-series data interacts with textual reports, visual data, or other information modalities. Future work will incorporate cross-modal corpora to enable more comprehensive information retrieval.

\section{Ethical Statement}
\label{append:ethical_statement}
Our framework and its contents are provided for academic and educational purposes only. None of the material constitutes financial, legal, or investment advice. No warranties, express or implied, are offered regarding the accuracy, completeness, or utility of the content. The authors and contributors are not responsible for any errors, omissions, or any consequences arising from the use of the information herein. Users should exercise their own judgment and consult professionals before making any financial, legal, or investment decisions. The use of the software and information contained in this repository is entirely at the user's own risk.

FinSrag and all its components are licensed under MIT.

\section{Task Definition Details}
\label{append:task_definition}
\textbf{Rise/Fall Threshold Settings.}
To classify daily movements as rise or fall, we first calculate returns $R_{t}$, which represents the percentage change in the closing price over consecutive days.
\begin{equation}
    \scalebox{0.9}{$
        \begin{aligned}
            R_{t} 
            &= \frac{\text{adj\_close$_d$} - \text{adj\_close$_{d-1}$}}{\text{adj\_close$_{d-1}$}} * 100
        \end{aligned}
    $}
\end{equation}
Following~\citet{yoo2021accurate} and ~\citet{soun2022accurate}, we classify the movement as rise if return $R_{t}$ exceeds 0.55, fall if it is below -0.5. In line with previous experimental settings~\citep{xie2023pixiu,xie2024finben}, we do not evaluate freeze cases as queries. However, we include sequences with $R_{t} \in [-0.5, 0.55]$ as freeze candidates in the candidate pool, ensuring a diverse and comprehensive set of historical data for context.
\begin{equation}
    \scalebox{0.9}{$
        \begin{aligned}
            M_{q,d} &= 
            \begin{cases} 
                \text{rise}, & R_{t} > 0.55 \\
                \text{fall}, & R_{t} < -0.5 \\
                \text{freeze}, &  -0.5 \leq R_{t} \leq 0.55
            \end{cases}
        \end{aligned}
    $}
\end{equation}

\textbf{Rationale for Unbalanced Thresholds.}
The asymmetrical thresholds for classifying daily movements (0.55 for rises and -0.5 for falls) reflect the inherent dynamics of stock market behavior. Stock prices typically rise gradually due to sustained investor optimism but fall sharply during panic selling or profit-taking. The stricter rise threshold prevents minor upward fluctuations from being misclassified as significant increases, while the more lenient fall threshold ensures meaningful downward trends are captured. This approach aligns with market realities, improving the reliability of movement classifications.

\section{Related Work}

\subsection{Stock Movement Prediction}
\textbf{Non-LLM Methods.}
Traditional approaches to stock movement prediction have focused on various aspects of financial data. One prominent category of methods analyzes stock price sequences and their corresponding technical indicators~\citep{qin2017dual,feng2018enhancing} to identify patterns in historical data for predicting future movements. However, due to the complexity of factors influencing stock prices, subsequent methods have incorporated additional contextual information, such as news articles~\citep{ding2015deep,liu2018hierarchical}  or social media posts~\citep{xu2018stock,wu2018hybrid}. Despite efforts, these methods are highly susceptible to noise and struggle to analyze the vast and diverse nature of financial information effectively.

\noindent
\textbf{LLM-based Methods.}
Recent studies have explored using LLMs for financial prediction tasks, either by fine-tuning open-source models or prompting advanced models like GPT-4. However, even state-of-the-art models, including GPT-4, have achieved only random-guessing-level accuracy in stock movement prediction~\citep{xie2023pixiu,xie2024finben,xie2023wall}. This highlights the inherent challenges in identifying and analyzing meaningful patterns in a domain as volatile and multifaceted as stock market prediction.

\subsection{Time-series forecasting with LLMs}
\textbf{Non-retrieval Methods.}
To enhance the performance of LLMs in time-series forecasting, existing methods have primarily focused on aligning temporal and textual data, either by transforming time-series into textual formats or by encoding both modalities into a unified vector space.
For instance, a study~\citep{jin2023time} reprograms time-series data into textual representations suitable for LLMs, enhancing prediction accuracy via declarative prompts. Similarly, some other studies~\citep{yu2023temporal,liu2024timecma} explore cross-modal alignment: the former applies LLMs to financial forecasting by integrating stock prices with news data, while the latter introduces a cross-modality framework to align time-series with text for improved predictive performance. 
While these advancements improve alignment, existing methods often struggle to process large-scale time-series data comprehensively due to LLM input limitations. As a result, effectively leveraging extensive historical data remains a challenge. This highlights the need for retrieval-augmented methods, a gap our approach directly addresses.

\noindent
\textbf{RAG Methods.}
With the rapid advancement of RAG techniques in various applications of LLMs, recent research has begun exploring RAG for time-series forecasting. For instance, TimeRAG~\citep{yang2024timerag} integrates RAG into time-series forecasting by combining Dynamic Time Warping (DTW) with LLMs to improve prediction accuracy. However, relying solely on numeric similarity is insufficient for financial time-series forecasting, as it fails to capture deeper semantic relationships. This underscores the need for a more targeted retrieval framework tailored to the complexities of financial data, a challenge our framework effectively addresses.

\section{Dataset Construction Details}
\subsection{Dataset Partition}
\label{append:dataset_partition}
To construct our datasets, we select high-trade-volume U.S. stocks across three periods: 2014-2015, 2017-2018, and 2022-2023. The first two periods align with two benchmark datasets: for 2014-2015, we use the same stocks as the ACL18 dataset~\citep{xu2018stock}, and for 2017-2018, we use the same stocks as the BIGDATA22 dataset~\citep{soun2022accurate}. To incorporate more recent data, we manually retrieve high-trade-volume stocks from 2022 to 2023 to create the STOCK23 dataset. This ensures our training and evaluation reflect recent market conditions and provide a robust benchmark for stock movement prediction.

To maintain temporal integrity and prevent information leakage, we implement stock-wise partitioning rather than temporal splitting of the dataset. This approach addresses a critical characteristic of our data where many high-trade-volume stocks (e.g., Google (GOOG)) appear across multiple years (2014-2015 and 2019-2020). 

Table \ref{tab:dataset_stock_info} details the complete stock lists and their corresponding sectors for all subsets, demonstrating two key properties of our partitioning: first, the complete absence of overlap between training and test sets, and second, comprehensive sector coverage in both sets. The alphabetical partitioning serves as a neutral, systematic approach that is uncorrelated with any financial or temporal characteristics while achieving our core objectives of clean separation and representative sampling. This method guarantees that the model encounters truly novel stocks during testing without any temporal leakage, while maintaining balanced sector representation across all partitions.

\begin{table}[ht]
\centering
\footnotesize
\renewcommand{\arraystretch}{1.2}
\caption{Stock information in datasets.}
\label{tab:dataset_stock_info}
\scalebox{0.85}{
\begin{tabular}{p{1.5cm}ccc}
\hline
\multicolumn{1}{c}{\textbf{Dataset}} & 
\multicolumn{1}{c}{\textbf{Split (amount)}} & 
\multicolumn{1}{c}{\textbf{Stock List}} & 
\multicolumn{1}{c}{\textbf{Sector Count}} \\
\hline
&  \multicolumn{1}{c}{train (33)} 
&  \multicolumn{1}{c}{\raisebox{-0.5ex}{
\makecell{
\fontsize{7}{11}\selectfont ABBV, AEP, AMGN, AMZN, \\
\fontsize{7}{11}\selectfont BA, BABA, BAC, BCH, BHP,\\
\fontsize{7}{11}\selectfont BP, BRK-A, BSAC, BUD, C, \\
\fontsize{7}{11}\selectfont CAT, CHTR, CMCSA, CODI,\\
\fontsize{7}{11}\selectfont CSCO, CVX, D, DHR, DIS,\\
\fontsize{7}{11}\selectfont DUK, EXC, GD, GE, GOOG,\\
\fontsize{7}{11}\selectfont HD, HON, HSBC, IEP, INTC
}}} & 
\multicolumn{1}{c}{\raisebox{-0.5ex}{
\makecell{ 
basic-materials (1), communication-services (4),\\
consumer-cyclical (3), consumer-defensive (1),\\
energy (3), financial-services (6), healthcare (3),\\
industrials (6), technology (2), utilities (4) 
}}} \\ \\
\multicolumn{1}{c}{ACL18}
&  \multicolumn{1}{c}{valid (5)} 
&  \multicolumn{1}{c}{\raisebox{-0.5ex}{
\makecell{
\fontsize{7}{11}\selectfont JNJ, JPM, KO, LMT, MA
}}} & 
\multicolumn{1}{c}{\raisebox{-0.5ex}{
\makecell{ 
consumer-defensive (1), financial-services (2),\\
healthcare (1), industrials (1)
}}} \\ \\
&  \multicolumn{1}{c}{test (33)} 
&  \multicolumn{1}{c}{\raisebox{-0.5ex}{
\makecell{
\fontsize{7}{11}\selectfont MCD, MDT, MMM, MO, MRK, \\
\fontsize{7}{11}\selectfont MSFT, NEE, NGG, NVS, ORCL,\\
\fontsize{7}{11}\selectfont PCG, PEP, PFE, PG, PM, PPL,\\
\fontsize{7}{11}\selectfont REX, SLB, SNY, SO, SPLP, \\
\fontsize{7}{11}\selectfont SRE, T, TM, TSM, UL, UNH, \\
\fontsize{7}{11}\selectfont UPS, V, VZ, WFC, WMT, XOM
}}} & 
\multicolumn{1}{c}{\raisebox{-0.5ex}{
\makecell{ 
basic-materials (1), communication-services (2),\\
consumer-cyclical (2), consumer-defensive (6),\\
energy (2), financial-services (2), healthcare (6),\\
industrials (3), technology (3), utilities (6)
}}} \\
\hline
&  \multicolumn{1}{c}{train (22)} 
&  \multicolumn{1}{c}{\raisebox{-0.5ex}{
\makecell{
\fontsize{7}{11}\selectfont AEP, AMGN, AMZN, BA, \\
\fontsize{7}{11}\selectfont BAC, C, CAT, CODI, CSCO,\\
\fontsize{7}{11}\selectfont CVX, D, DIS, DUK, EXC, \\
\fontsize{7}{11}\selectfont GD, GE, GMRE, GOOG, HD, \\
\fontsize{7}{11}\selectfont HON, INTC, JNJ
}}} & 
\multicolumn{1}{c}{\raisebox{-0.5ex}{
\makecell{ 
communication-services (2), consumer-cyclical (2), \\
energy (1), financial-services (2), healthcare (2) \\
industrials	(6), real-estate (1), technology (2), utilities (4)
}}} \\ \\
\multicolumn{1}{c}{BIGDATA22}
&  \multicolumn{1}{c}{valid (3)} 
&  \multicolumn{1}{c}{\raisebox{-0.5ex}{
\makecell{
\fontsize{7}{11}\selectfont JPM, KO, LMT
}}} & 
\multicolumn{1}{c}{\raisebox{-0.5ex}{
\makecell{ 
financial-services (1),
consumer-defensive (1), \\
industrials (1)
}}} \\ \\
&  \multicolumn{1}{c}{test (22)} 
&  \multicolumn{1}{c}{\raisebox{-0.5ex}{
\makecell{
\fontsize{7}{11}\selectfont MA, MCD, MDT, MMM, MO,\\
\fontsize{7}{11}\selectfont MRK, MSFT, NEE, ORCL, PCG,\\
\fontsize{7}{11}\selectfont PM, PPL, REX,  SO, SRE,\\
\fontsize{7}{11}\selectfont T, UPS, V, VZ, WFC, \\
\fontsize{7}{11}\selectfont WMT, XOM
}}} & 
\multicolumn{1}{c}{\raisebox{-0.5ex}{
\makecell{ 
basic-materials (1), communication-services (2),\\
consumer-cyclical (1), consumer-defensive (3),\\
energy (1), financial-services (3), healthcare (2),\\
industrials (2), technology (2), utilities (6)
}}} \\
\hline
&  \multicolumn{1}{c}{train (24)} 
&  \multicolumn{1}{c}{\raisebox{-0.5ex}{
\makecell{
\fontsize{7}{11}\selectfont ADBE, ADSK, AMD, AVGO, \\
\fontsize{7}{11}\selectfont COIN, COST, DELL, DXCM, ELV,\\
\fontsize{7}{11}\selectfont ETSY, FOX, HOOD, HPQ, IBM, \\
\fontsize{7}{11}\selectfont INTU, JBL, JNJ, KKR, KMX, \\
\fontsize{7}{11}\selectfont LDOS, LLY, LOW, LULU, LUV
}}} & 
\multicolumn{1}{c}{\raisebox{-0.5ex}{
\makecell{ 
communication-services (1), consumer-cyclical (4),\\
consumer-defensive (1), financial-services (3),\\
healthcare (4), industrials (1), technology (10)
}}} \\ \\
\multicolumn{1}{c}{STOCK23}
&  \multicolumn{1}{c}{valid (3)} 
&  \multicolumn{1}{c}{\raisebox{-0.5ex}{
\makecell{
\fontsize{7}{11}\selectfont MARA, META, MNST
}}} & 
\multicolumn{1}{c}{\raisebox{-0.5ex}{
\makecell{ 
financial-services (1),
communication-services (1), \\
consumer-defensive (1)
}}} \\ \\
&  \multicolumn{1}{c}{test (24)} 
&  \multicolumn{1}{c}{\raisebox{-0.5ex}{
\makecell{
\fontsize{7}{11}\selectfont MRVL, MS, MSCI, NDAQ,\\
\fontsize{7}{11}\selectfont NKE, NVDA, ORLY, PANW, PEAK,\\
\fontsize{7}{11}\selectfont PLTR, PNC, PNW, RY, SMCI,\\
\fontsize{7}{11}\selectfont SNOW, SNPS, T, TGT, TSLA,\\
\fontsize{7}{11}\selectfont UBER, ULTA, V, WBA, WDAY
}}} & 
\multicolumn{1}{c}{\raisebox{-0.5ex}{
\makecell{ 
communication-services (1), consumer-cyclical (4),\\
consumer-defensive (1), financial-services (6),\\
healthcare (2), technology (9), utilities (1)
}}} \\
\hline
\end{tabular}
}
\end{table}

\subsection{Financial Indicators}
\textbf{Financial Indicators Selection}
\label{append:mi_score}
We compute the MI scores for each indicator using nonparametric estimation methods provided by \textit{mutual\_info\_regression} from the Scikit-learn library. 
For each query with retrieved candidates, we convert the movements of retrieved sequences and LLM's prediction to numerical representations.
\begin{equation*}
    \scalebox{0.9}{
        $\begin{aligned}
            {M_{c_j}, M_{q}} &= 
            \begin{cases} 
                \text{1}, & \text{rise} \\
                \text{0}, & \text{freeze} \\
                \text{-1}, & \text{fall}
            \end{cases}
        \end{aligned}$
    }
\end{equation*}
Then, we calculate the average value of the $M_{c_i}$ from the five retrieved sequences to represent the RAG-provided result, and the final prediction of the query is recorded as another numerical value:
\begin{equation*}
    \scalebox{0.9}{
        $\begin{aligned}
            x_i &= \frac{1}{5}\sum_{j=0}^4 M_{c_j}, \\
            y_i &= M_{q_i}.
        \end{aligned}$
    }
\end{equation*}
Lastly, we adopt the Pearson correlation coefficient to compute the correlation:
\begin{equation*}
    \scalebox{0.9}{
        $\begin{aligned}
            r = \frac{\sum (x_i - \bar{x})(y_i - \bar{y})}{\sqrt{\sum (x_i - \bar{x})^2 \sum (y_i - \bar{y})^2}}.
        \end{aligned}$
    }
\end{equation*}

\begin{figure}[ht]
    \centering
    \includegraphics[width=0.4\linewidth]{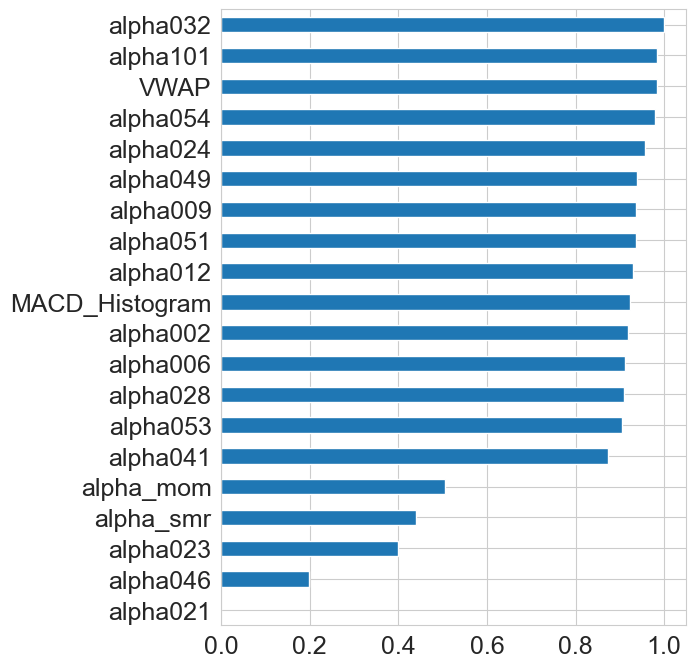}
    \caption{Mutual Information(MI) scores between indicators and forward return.}
    \label{fig:mi}
\end{figure}

Then we normalize them into range $(0,1)$. These scores provide insight into the strength of the dependency between each financial indicator and the forward return. Higher MI scores indicate stronger relationships, highlighting which indicators are most predictive of future price movements. We select the top-18 alpha indicators with the highest MI scores in our candidates, as is shown in Figure \ref{fig:mi}. 


\clearpage
\textbf{Indicators Descriptions} We provided a brief description of each indicator in Table \ref{tab:indicator_description}.
\label{append:indicator_description}

\begin{table}[h]
\tiny
\centering
\renewcommand{\arraystretch}{1.3}
\caption{Brief descriptions of indicators.}
\label{tab:indicator_description}
\begin{tabular}{m{0.6cm}m{4cm}m{8cm}}
\toprule
\textbf{Group} & \textbf{Indicator} & \textbf{Description} \\
\hline
\multicolumn{3}{c}{\textit{basic price indicators}} \\
\hline
\multirow{5}{*}{1} & open (opening price) \cite{CoinAPI2023}
& reflecting the stock’s value at the start of the trading session. 
\\
& high, low (highest/lowest price) \cite{CoinAPI2023}
& capture the price range during the day, offering insights into volatility.
\\
& close (close price) \cite{CoinAPI2023}
& often the most important for technical analysis, represents the final trading price for the day. 
\\
& adj\_close (adjusted close price) \cite{Ganti2020}
& accounts for corporate actions like dividends or stock splits, providing a more accurate long-term price history.
\\
 & volume \cite{CoinAPI2023}
& indicating the level of trading activity and movement, typically the change in price from the previous day, reflects short-term price fluctuations. 
\\
\hline
\multicolumn{3}{c}{\textit{devived financial indicators}} \\ \hline
\multirow{3}{*}{2} & movement \cite{Tardi2022, Jegadeesh1990, Jegadeesh1993}
& represents the change in stock price between consecutive trading days. 
\\
&returns \cite{Tardi2022, Jegadeesh1990, Jegadeesh1993}
& represents the percentage change in the closing price over consecutive days. 
\\
&VWAP (Volume Weighted Average Price) \cite{Mitchell2020}
& calculates the average price of a stock weighted by trading volume, providing a benchmark for institutional investors.  
\\ \hline
\multirow{2}{*}{3} &macd\_histogram (values) \cite{Anghel2015}
& measures the difference between a stock's MACD line and its signal line, helping identify trends and momentum. 
\\
&macd\_crossover (signals) \cite{Anghel2015}
& indicates a buy or sell signal when the MACD line crosses above or below the signal line. 
\\ \hline
\multirow{2}{*}{4} &bollinger\_bands (signals) \cite{Bollinger2001}
& indicates whether the adjusted close price exceeds the upper or lower price limits around a moving average, serving as a measure of price volatility. 
\\
& exceeding\_upper/lower (values) \cite{Bollinger2001}
& the value of the adjusted close price exceeds the upper or lower price limits around a moving average.
\\ \hline
\multirow{2}{*}{5} &overbought\_and\_oversold\_conditions \cite{Wilder1978}
& classify the market state into "overbought" or "oversold" areas based on predefined thresholds for \( K \), \( D \), and \( J \).  
\\
&kdj\_crossover (signals) \cite{wu2015technical}
& identify "bullish" or "bearish" signals based on the relationship between \( K \) and \( D \).
\\ \hline
\multirow{9}{*}{6} &alpha\_smr \cite{kakushadze2016101}
& measures short-term reversal patterns by capturing stocks' tendency to revert after abrupt price movements.
\\
&alpha\_mom \cite{kakushadze2016101}
& tracks momentum effects by identifying stocks with strong recent performance likely to continue trending.
\\
&alpha\_009 \cite{kakushadze2016101}
& uses recent price changes to classify trends based on their range within a 5-day window.
\\
&alpha\_012 \cite{kakushadze2016101}
& relates volume change direction to the negative change in price.
\\
&alpha\_041 \cite{kakushadze2016101}
& measures the difference between price range midpoint and VWAP.
\\
&alpha\_049 \cite{kakushadze2016101}
& detects strong downtrends or reversals in lagged price movements.
\\
&alpha\_051 \cite{kakushadze2016101}
& quantifies mean-reversion behavior over medium horizons, typically exploiting price corrections after sustained moves.
\\
&alpha\_054 \cite{kakushadze2016101}
& combines momentum and volume signals to detect stocks with improving trends supported by trading activity.
\\
&alpha\_101 \cite{kakushadze2016101}
& normalizes price change by daily price range.
\\ \hline
\multirow{6}{*}{7} &alpha\_002 \cite{kakushadze2016101}
& measures the negative 6-day correlation between ranked changes in logarithmic volume and ranked price movements.
\\
&alpha\_006 \cite{kakushadze2016101}
& computes the negative correlation between opening price and volume over a 10-day window.
\\
&alpha\_024 \cite{kakushadze2016101}
& compares long-term and short-term price movements to identify breakouts or retracements.
\\
&alpha\_028 \cite{kakushadze2016101}
& scales the relationship between average price, low price correlation, and close price.
\\
&alpha\_032 \cite{kakushadze2016101}
& combines deviations from a 7-day moving average and long-term correlation between VWAP and lagged close.
\\
&alpha\_053 \cite{kakushadze2016101}
&  identifies stocks exhibiting strong recent performance (typically over 3-6 months) while controlling for volatility, aiming to capture persistent trends while mitigating risk from erratic price swings.
\\ \hline
\multirow{3}{*}{8} &alpha\_021 \cite{kakushadze2016101}
& identifies trends based on short- and long-term price averages and volume conditions.
\\
&alpha\_023 \cite{kakushadze2016101}
& tracks changes in high prices if they exceed their 20-day average.
\\
&alpha\_046 \cite{kakushadze2016101}
& analyzes trends in lagged and current 10-day price changes.
\\ \bottomrule
\end{tabular}
\end{table}

\subsection{Sequences and Serialization}
\label{sec:datastore_example}
An example in our datastore is shown below.
\begin{lstlisting}[language=Java]
{
    "stock_name":"ABBV",
    "query_date":"2014-06-05",
    "movement":"rise",
    "open":54.549999,
    "high":55.32,
    "low":54.360001,
    "close":55.299999,
    "adj_close":36.980961,
    "volume":4847300,
    "macd_histogram":0.0621141602,
    "macd_crossover":null,
    "bollinger_bands":null,
    "exceeding_upper":null,
    "exceeding_lower":null,
    "overbought_and_oversold_conditions":null,
    "kdj_crossover":null,
    "returns":0.0131913094,
    "VWAP":54.9933333333,
    "alpha_smr":0.000549858,
    "alpha_mom":0.0121661224,
    "alpha_002":null,
    "alpha_006":null,
    "alpha_009":-0.481476,
    "alpha_012":-0.481476,
    "alpha_021":1,
    "alpha_023":0.0,
    "alpha_024":-0.769035,
    "alpha_028":null,
    "alpha_032":null,
    "alpha_041":-0.1554335256,
    "alpha_046":-0.481476,
    "alpha_049":-0.481476,
    "alpha_051":-0.481476,
    "alpha_053":null,
    "alpha_054":126.4264005641,
    "alpha_101":-18.282056485
}
\end{lstlisting}

\section{Experimental Settings}
\subsection{StockLLM-1B-Instruct Backbone}
\label{append:finetuning_details}
To activate the LLM's inherent knowledge and ensure instruction-following capabilities, we fine-tune a 1B parameter LLM (LLaMA 3.2-1B-Instruct)~\citep{grattafiori2024llama3herdmodels} using the LoRA technique for efficient low-rank adaptation, resulting in StockLLM-1B-Instruct. By intentionally using a smaller backbone model, we establish a more challenging experimental setup, ensuring that performance improvements are attributable to FinSeer's retrieval capabilities rather than the LLM's capacity. More details of fine-tuning are illustrated in Appendix~\ref{append:finetuning_details}.

The fine-tuning process is implemented using the LlamaFactory framework~\citep{zheng2024llamafactory}, with the following configuration: a learning rate of 5e-5, a cosine scheduler, gradient accumulation over 8 steps, mixed-precision (fp16) training, and 5 epochs of training with regular evaluation to log metrics and select the best-performing checkpoint. This setup demonstrates the robustness of our framework in extracting meaningful insights even under constrained model size, highlighting the effectiveness of FinSeer in enhancing financial forecasting tasks. The fine-tuning prompt is shown in Figure~\ref{fig:prompt_rag}, with candidate sequences randomly retrieved from the dataset. The fine-tuning step is to train the StockLLM-1B-Instruct to follow instructions, without additional steps for retrieving certain candidates.

\subsection{Experimental Settings}
\label{append:baseline_settings}
\textbf{Baselines.} The introduction of retrieval baselines is as follows:

\begin{itemize}
    \item Random retrieval: The candidate range is aligned with the "Candidate Pool Settings" part. We use the random.sample() function in Python to retrieve random candidates from the candidate pool. 
    \item DTW (Dynamic Time Warping) distance: The formula measures the similarity between two temporal sequences by aligning them non-linearly in time, minimizing the cumulative warping cost.  $D_{\text{DTW}}(X, Y) = \min_{\pi \in \mathcal{P}} \sqrt{\sum_{(i,j) \in \pi} (x_i - y_j)^2}$, where \(X, Y\) are sequences, \(\pi\) is a warping path, and \(\mathcal{P}\) is the set of all possible paths.
    \item Instructor-large~\citep{SuSKWHOYSZ023}: A 335M instruction-finetuned text embedder that encodes sequences into 768-dimensional tensors. 
    \item  UAE-large-v1~\citep{li2023angle}: A 335M ANGLE-optimized text embedding model that encodes sequences into 1024-dimensional tensors. 
    \item E5-mistral-7b-instruct~\citep{wang2023improving}: A 7111M embedder initialized from Mistral-7B-v0.1~\citep{jiang2023mistral} and fine-tuned on multilingual datasets, encoding sequences into 4096-di-mensional tensors.  
    \item BGE-large-en-v1.5~\citep{bge_embedding}: A 335M general embedder pre-trained with RetroMAE~\citep{RetroMAE}, encoding sequences into 1024-dimensional tensors.
    \item LLM-Embedder~\citep{zhang2023retrieve}: A 109M embedder fine-tuned from BGE-large-en-v1.5, designed as a unified embedding model to support diverse retrieval augmentation needs for LLMs. It encodes sequences into 768-dimensional tensors. 

\end{itemize}

\begin{figure}[ht]
    \centering
    \includegraphics[width=0.8\linewidth]{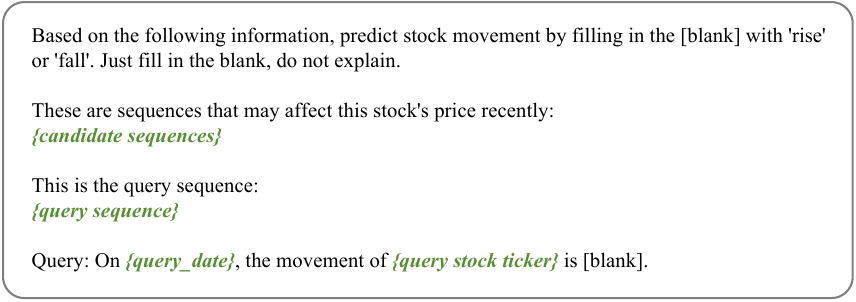}
    \caption{The prompt template for StockLLM-1B-Instruct fine-tuning and RAG methods evaluation.}
    \label{fig:prompt_rag}
\end{figure}

\begin{figure}[hb]
    \centering
    \includegraphics[width=0.82\linewidth]{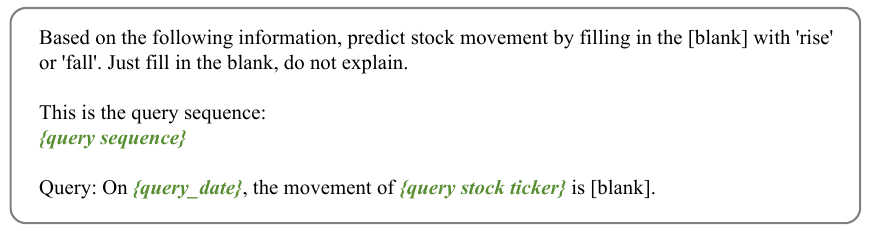}
    \caption{The prompt template for bare StockLLM-1B-Instruct evaluation.}
    \label{fig:prompt_bare_llm}
\end{figure}

\textbf{Prompts.} The prompt for bare StockLLM-1B-Instruct is shown in Figure~\ref{fig:prompt_bare_llm}, and the prompt for retrieval methods is shown in Figure~\ref{fig:prompt_rag}. 

\textbf{Hyperparameters.} For each model, we employ default parameter settings, utilizing official
models for open-source LLMs obtained from Hugging Face. These testing procedures take place on a computational infrastructure consisting of one NVIDIA A800 Tensor Core GPU, equipped with 80GB of memory.

\textbf{Error Bars.}
To account for potential variability, we conducted three randomized retrieval trials and report averaged performance metrics. For similarity-based retrieval methods, since both queries and candidates are encoded as fixed vector representations, the retrieved candidates remain deterministic across runs. To verify this stability, we performed three identical inference passes through StockLLM, observing minimal variance in prediction outcomes.

\newpage
\section{Addtional Experimental Results}

\subsection{Candidate Movement Correlation}
\label{append:candidate_movement_relation}
In this section, we investigate whether StockLLM-1B-Instruct relies on the movement trends of retrieved sequences to make predictions. To this end, we compute the correlation between the movements of retrieved sequences and the LLM's generated results.

The results in Table \ref{tab:correlation_movement_count} reveal no significant correlation between the movement direction (rise or fall) of the retrieved candidate sequences and the final predictions of StockLLM. This finding indicates that StockLLM-1B-Instruct does not simply mirror the movement trends of the retrieved sequences but instead analyzes their specific content to infer the query's movement direction. This ability highlights the LLM's capacity to extract meaningful insights from complex time-series data.
\begin{table}[ht]
    \centering
    \caption{Correlation between the movements of retrieved sequences and LLM's generated results.}
    \label{tab:correlation_movement_count}
    \scalebox{0.7}{
    \begin{tabular}{cccc}
    \toprule
    \text{Dataset} & \text{RAG Model} & \text{Correlation} & \text{P-Value} \\ \hline
         \multirow{6}{*}{ACL18} & Instructor & -0.098 & 0.000 \\
                & UAE        & -0.068 & 0.000 \\
                & E5         & -0.108 & 0.000 \\
                & BGE        &  0.175 & 0.000 \\
                & LLM Embedder & 0.014 & 0.457 \\
                & FinSeer    &  0.054 & 0.004 \\
\midrule
\multirow{6}{*}{BigData22} & Instructor & -0.072 & 0.000 \\
                   & UAE        &  0.009 & 0.633 \\
                   & E5         & -0.040 & 0.032 \\
                   & BGE        &  0.093 & 0.000 \\
                   & LLM Embedder & 0.060 & 0.001 \\
                   & FinSeer    &  0.077 & 0.000 \\
\midrule
\multirow{6}{*}{STOCK23} & Instructor & -0.114 & 0.000 \\
                 & UAE        & -0.093 & 0.000 \\
                 & E5         & -0.064 & 0.000 \\
                 & BGE        &  0.131 & 0.000 \\
                 & LLM Embedder & 0.015 & 0.350 \\
                 & FinSeer    &  0.032 & 0.041 \\
                 \bottomrule
    \end{tabular}}
\end{table}

This observation underscores the critical role of retrieval quality in RAG models, as the relevance and informativeness of retrieved sequences directly influence prediction performance. Notably, StockLLM-1B-Instruct is a relatively small 1B parameter LLM, making this a particularly challenging setting. Despite this, our time-series-tailored RAG model, FinSeer, successfully enhances StockLLM's performance by retrieving rich and relevant time-sensitive information. This demonstrates the effectiveness of our approach in improving prediction accuracy and showcases its potential for financial time-series forecasting.

\subsection{Indicator Sequence Visualization}
\label{append:visualization}
\begin{figure}[ht]
    \centering
    \includegraphics[width=\linewidth]{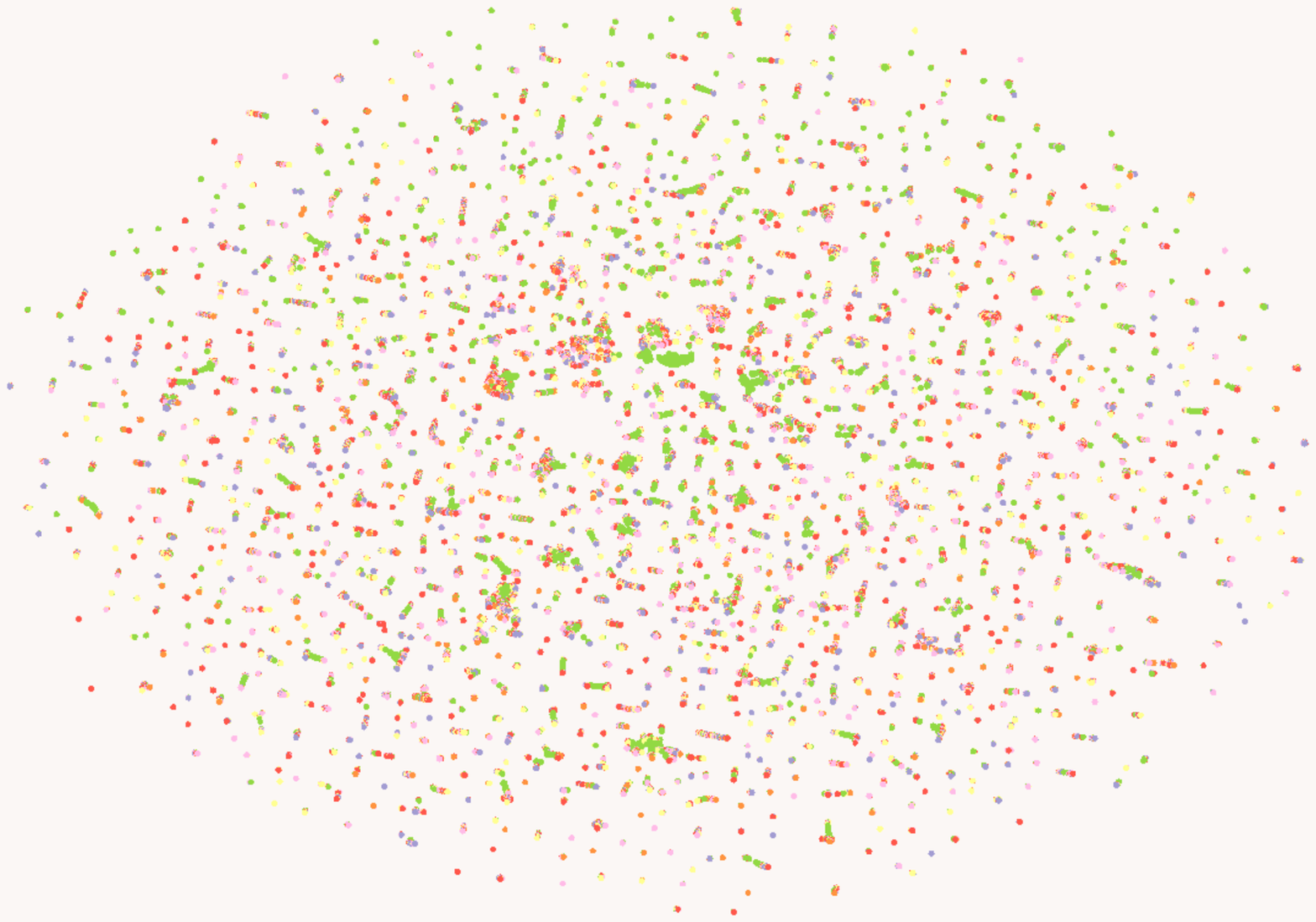}
    \caption{Visualization of indicator sequence embeddings on ACL18 dataset.}
    \label{fig:visualization}
\end{figure}

To intuitively examine how FinSeer embeds indicator sequences, we use LargeVis~\citep{tang2016visualizing} to reduce the embeddings to two dimensions. In Figure \ref{fig:visualization}, we visualize\footnote{https://medviz.org/app/} the vector space of nine indicators, including adjusted close price, open price, high price, low price, close price, volume, movement, returns, and WAP. The results show that the green points, representing movement information, are well-clustered, indicating that FinSeer effectively captures meaningful patterns. However, this also suggests that deeper relationships between stock movements and indicators remain to be explored, highlighting the potential for further research in this area.

\subsection{Results of StockLLM-8B-Instruct}
\label{append:result-8b}
We additionally trained StockLLM-8B-Instruct based on LLaMA-3.1-8B-Instruct to evaluate FinSeer's performance on larger LLMs. As shown in Table \ref{tab:main-result-8b}, the results demonstrate that FinSeer can consistently enhance LLM performance regardless of the backbone model's inherent capability for stock movement prediction. Compared to the 1B version, we observe greater performance variability in the bare LLM, while different retrieval methods maintain similar relative impacts on StockLLM's performance. Notably, random retrieval and other retrieval methods continue to show inconsistent effects, sometimes improving but other times degrading performance.. FinSeer not only outperforms all retrieval-based approaches but also delivers stable performance enhancements to StockLLM, validating the effectiveness of our RAG framework.

\begin{table}[ht]
\small
\centering
\renewcommand{\arraystretch}{1.3}
\caption{Results of stock movement predictions using StockLLM-8B-Instruct and retrieval models.}
\label{tab:main-result-8b}
\scalebox{0.85}{
\begin{tabular}{ccccccccc}
\hline
{Retrieving Methods}
& \multicolumn{2}{c}{ACL18} 
&  
& \multicolumn{2}{c}{BIGDATA22} 
&  
& \multicolumn{2}{c}{STOCK23} 
\\ 
\cline{2-3} 
\cline{5-6} 
\cline{8-9} 
(+ StockLLM-8B-Instruct)
& ACC & MCC &  
& ACC & MCC & 
& ACC & MCC 
\\ \hline
w/o Retrieval   
& 0.479 & -0.041 &
& 0.446 & -0.108 & 
& 0.510 & 0.020    \\
Random Retrieval 
& 0.509 & 0.028 &
& 0.450 & -0.149 & 
& 0.495 & -0.012 \\
DTW
& 0.509 & 0.020 &
& 0.457 & -0.087 &
& 0.496 & -0.008\\
Instructor     
& 0.497 & -0.005 &
& 0.457 & -0.087 &
& 0.506 & 0.011 \\
UAE             
& 0.509 & 0.019 &
& 0.465 & -0.081 &
& 0.509 & 0.018 \\
E5              
& 0.484 & -0.031 &
& 0.453 & -0.099 &
& 0.498 & -0.005 \\
BGE             
& 0.522 & 0.058 &
& 0.460 & -0.104 & 
& 0.507 & -0.001 \\
LLM Embedder    
& 0.486 & -0.034 &
& 0.454 & -0.123 & 
& 0.499 & -0.002 
\\ \hline
FinSeer        
& \textbf{0.554} & \textbf{0.122} & 
& \textbf{0.469} & \textbf{-0.065} & 
& \textbf{0.511} & \textbf{0.023} 
\\ \hline
\end{tabular}
}
\end{table}

\end{document}